\documentclass{Interspeech}

\interspeechcameraready

\title{Prompting Whisper for Improved Verbatim Transcription and End-to-end Miscue Detection}

\author[affiliation={1*}]{Griffin}{Dietz Smith}
\author[affiliation={1*}]{Dianna}{Yee}
\author[affiliation={1}]{Jennifer}{King Chen}
\author[affiliation={1}]{Leah}{Findlater}

\affiliation{Apple}{USA}

\email{\{griffind, dianna\_yee, jennkingchen, lfindlater\}@apple.com}

\usepackage{comment}

\usepackage{graphicx}
\usepackage{subfig}
\usepackage{booktabs} 
\usepackage{multirow}
\usepackage{dblfloatfix}
\usepackage{cite}

\begin{document}

\maketitle

\def\thefootnote{*}\footnotetext{These authors contributed equally to this work}\def\thefootnote{\arabic{footnote}}

\begin{abstract}
Identifying mistakes (i.e., miscues) made while reading aloud is commonly approached post-hoc by comparing automatic speech recognition (ASR) transcriptions to the target reading text. 
However, post-hoc methods perform poorly when ASR inaccurately transcribes verbatim speech. To improve on current methods for reading error annotation, we propose a novel end-to-end architecture that incorporates the target reading text via prompting and is trained for both improved verbatim transcription and direct miscue detection. Our contributions include: first, demonstrating that incorporating reading text through prompting benefits verbatim transcription performance over fine-tuning, and second, showing that it is feasible to augment speech recognition tasks for end-to-end miscue detection. We conducted two case studies---children's read-aloud and adult atypical speech---and found that our proposed strategies improve verbatim transcription and miscue detection compared to current state-of-the-art.
 
\end{abstract}

\section{Introduction}
The practice of annotating reading errors (i.e., miscues) in read-aloud text has applications in education, speech-language pathology, and speech data collection, and can be used to benchmark, track progress, or identify areas for intervention. However, annotating reading errors is traditionally a time-consuming manual process that involves determining what people say while reading aloud (i.e., verbatim transcription) and identifying miscues present in speech (i.e., miscue detection [MD]). 

Recently, researchers have begun exploring automated methods for reading annotation. Because miscues---and pronunciation scores in the neighboring space of computer-aided pronunciation training---are commonly classified post-hoc from automatic speech recognition (ASR) or phoneme recognition transcriptions \cite{duchateau2006flexible, gao2024reading, black2007automatic, proencca2018mispronunciation}, one might expect recent advances in ASR to improve reading annotation. However, modern large-scale ASR systems are not robust to speech that falls out of the distribution of typical training datasets, like children's or atypical speech that is often the target of reading annotation. Furthermore, ASR is typically trained to ignore dysfluencies that MD specifically aims to detect (e.g., filler words and repetitions), so it rarely captures a true verbatim. Efforts have been made to improve ASR through fine-tuning for such targeted groups~\cite{southwell, graave24_interspeech, sparsesharedlora, rolland24b_interspeech, lea2023stuttering, tobin2024singleasrmodelgeneralizes,tobin2024automatic}, but when reading annotation relies solely on ASR, incorrect transcriptions lead to inaccurate MD.

Critically, ASR models typically fail to account for the target reading text, despite that text containing rich contextual information for reading annotation tasks. Therefore, researchers often focus on improving verbatim transcription performance by attuning to the target reading text. Some approaches use task-specific language models constrained on the target reading text to improve recognition \cite{gothi2024dataset, mostow2012and} and apply post-hoc MD using additional information such as estimated goodness of pronunciations or ASR confidence values. These targeted language models can improve precision, but are not robust to data shifts in the target reading text, nor address challenges with recognizing misread or non-words not present in that target text. Recently researchers explored incorporating the target reading text through prompting of an off-the-shelf large-scale ASR system to improve verbatim transcription performance in Dutch, but performance of post-hoc calculated miscues did not surpass task-specific language models trained on manual transcriptions and target reading text~\cite{molenaar23_interspeech}. In the pronunciation training space, similar strategies have been explored where the target text is directly encoded into input embeddings and fused with audio input via attention mechanisms to improve phone recognition performance \cite{feng2020sed, fu2021full}, but this approach still relies on post-hoc calculation of transcripts and has not been applied to MD. 

Limited efforts have been made towards truly end-to-end (E2E) MD, where a model directly predicts miscues in addtion to verbatim, rather than relying on post-hoc calculation. One contribution proposes an E2E approach where the reading text and audio are fused through an attention mechanism with separate LSTM decoders for ASR, dysfluency detection, and classification ~\cite{venkatasubramaniam2023end}. However, dysfluencies constitute only a subset of miscue events and this approach has no ability to detect insertion miscues. Furthermore this work reported low macro-F1 scores for dysfluency detection on children's speech (\textless 0.45), and did not report verbatim transcription accuracy. 

To improve on current methods for reading annotation, we propose and evaluate a novel E2E approach tuned for both verbatim transcription and direct MD. This model considers target reading text information, transcribes how that speech was read, and identifies miscues. We conducted two case studies comparing our proposed approach to post-hoc calculation approaches. The first case study focuses on children's speech (age 5--9) that contains reading errors common in early education. The second case study focuses on adult atypical speech, where miscues may arise due to dysfluencies (e.g., stutters) and pronunciation differences. Our contributions include: 1) demonstrating that incorporating target reading text through prompting benefits verbatim transcription performance over direct fine-tuning and 2) showing it is feasible to augment ASR tasks for E2E reading annotation using our proposed design.
\section{Method}
\label{section:method}

\begin{figure}
    \includegraphics[width=\linewidth]{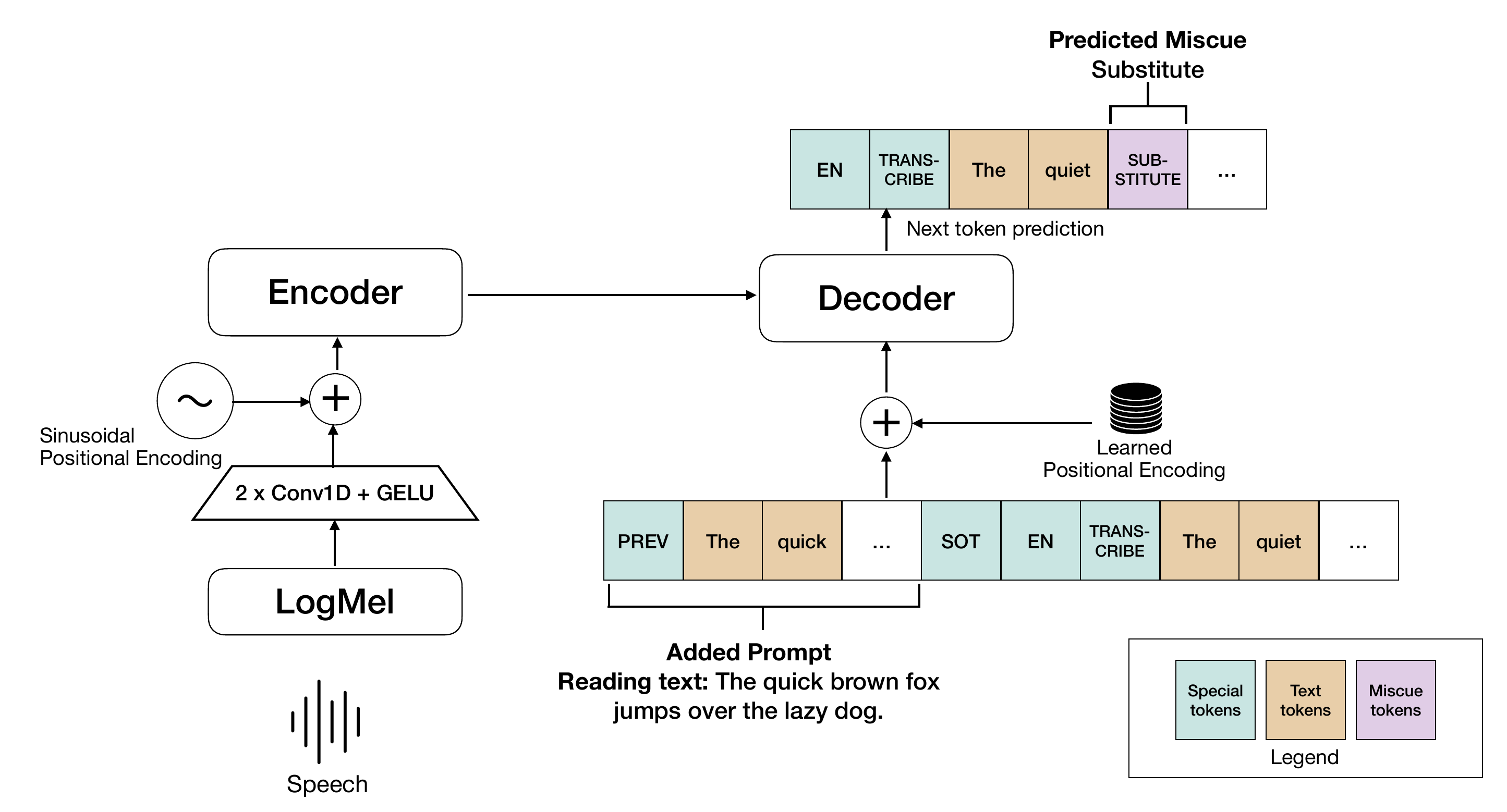}
    \caption{Proposed architecture for end-to-end miscue detection by augmenting Whisper to include miscue tokens and incorporate reading text via prompting.}
    \label{fig:miscue_architecture}
\end{figure}

We propose modifying the canonical ASR task for MD with two architectural changes. Our first change fine-tunes the pre-trained ASR while prompting with the target reading text. This prompting is a systematic way to incorporate additional relevant context from the reading text during speech recognition and should help improve both transcription performance and subsequently, post-hoc calculated miscues. 
Our second modification extends the first approach to modify ASR for E2E MD. As shown in Figure \ref{fig:miscue_architecture}, in addition to the aforementioned prompting strategy, we augment the tokenizer vocabulary to include annotations for miscue events. Therefore, the E2E system is trained to incorporate context from the target reading text to directly predict both verbatim transcriptions and miscue annotations. 

We apply our proposed strategies to Whisper~\cite{whisper-radford}, a large-scale speech foundation model trained on 680k hours of speech, to investigate the impact of prompting on verbatim speech transcription and the feasibility of an E2E approach towards MD.

\subsection{Incorporation of prompts}
\label{section:promptdesign}

To train Whisper with target reading text prompts as additional context for prediction, we tokenize the reading text and prepend it to the `Start of Token' (\textless\textsc{sot}\textgreater). To avoid training the model to predict the prompt itself, the computed loss considers only predicted verbatim transcriptions and miscue detections and the corresponding ground truth labels. Because Whisper has a maximum audio duration of 30 seconds and 448 tokens, all samples are clipped to be within limits.

\subsection{Incorporation of miscue types}
\label{section:miscuedesign}

\begin{figure}
    \includegraphics[width=\linewidth]{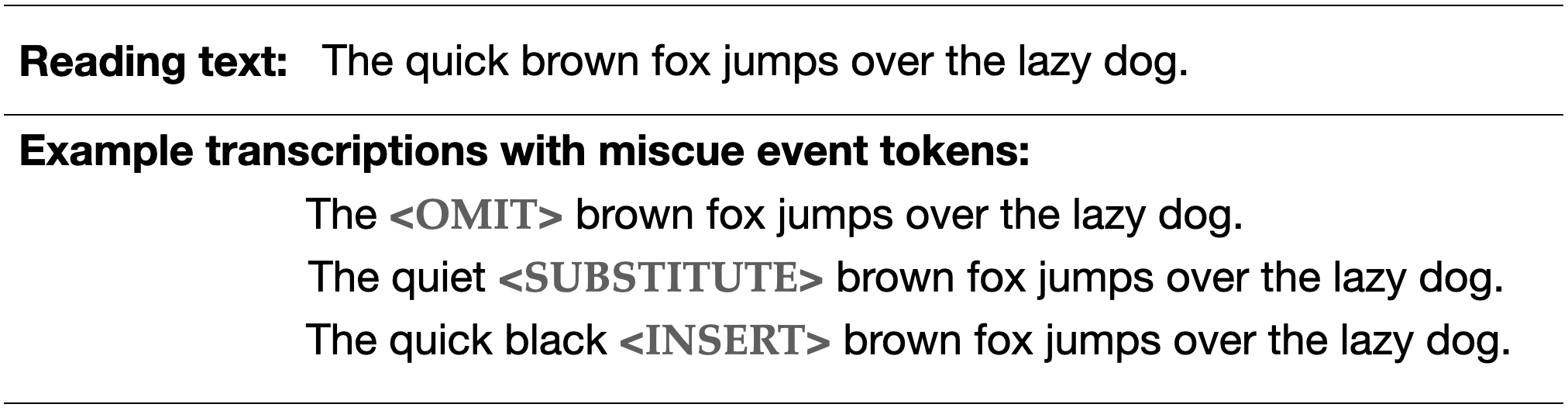}
    \caption{Example of a target reading text prompt and potential miscue events defined in Section \ref{section:method}.}
    \label{fig:miscue_examples}
\end{figure}

To incorporate different miscues types, we augment the vocabulary of the tokenizer to include tokens associated with miscue events: \{\textless\textsc{omit}\textgreater, \textless\textsc{substitute}\textgreater, \textless\textsc{insert}\textgreater\}. An \textless\textsc{omit}\textgreater\hspace{0.01mm} event occurs when the reader does not read a word, a \textless\textsc{substitute}\textgreater\hspace{0.01mm} event occurs when the reader reads a different word than specified, and an \textless\textsc{insert}\textgreater\hspace{0.01mm} event occurs when the reader inserts a word that is not specified. Examples of such miscue events are shown in Figure \ref{fig:miscue_examples} for clarity.

This framework is theoretically extensible to other speech events such as pauses or filler words. Due to the limited availability of such annotated data and a desire to compare with post-hoc approaches, we focus this work on miscue events that are computable with common string comparison operations. 

\section{Experiments}

\subsection{ASR model}
We ran experiments with various sizes of Whisper \{\texttt{tiny.en, small.en, medium.en}\} to better understand the effect of model size on the performance of the proposed approach. We use English-only models as all our datasets are in English.

\subsection{Data}
\label{subsec:data}

To examine our approach, we consider two case studies where speakers were tasked with reading short sentences aloud: typical children's and atypical adult speech.

In our first case study, we use a dataset, $X_\mathrm{c}$, which consists of over 124,000 utterances from 367 children aged 5 to 9. All utterances were sentences read from fictional texts.
These utterances contain reading miscues common in early education that naturally arise due to blending or tracking mistakes, lacking phonics knowledge, or incorrect guesses. Transcriptions of these utterances capture those miscues verbatim. 

Our second case study focuses on adults with atypical speech using data collected through the Speech Accessibility Project \cite{SAProject}. This dataset, denoted by $X_\mathrm{a}$, consists of over 180,000 read speech utterances from 616 speakers who have Parkinson's disease, ALS, cerebral palsy, or Down syndrome. We filtered for read speech where the target reading texts are sentences from fictional books or common voice assistant tasks. This dataset consists of miscues which may arise either naturally or due to dysfluencies such as whole or partial word repetitions; in this dataset as well, miscues are transcribed verbatim. 

\begin{table*}[t]
    \centering
    \resizebox{.8\linewidth}{!}{
    \begin{tabular}{llccccccc}
    \toprule
        \textsc{Data} & \textsc{Model} & \textsc{Naive} 
        & \multicolumn{2}{c}{\textsc{Untuned}}& 
        \multicolumn{2}{c}{\textsc{Tuned}} & \textsc{End-to-end} \\
        \cmidrule(lr){4-5}
        \cmidrule(lr){6-7}
        \cmidrule(lr){8-8}
          &   &   & \textsc{Unprompt} & \textsc{Prompt} & \textsc{Unprompt} & \textsc{Prompt} [ours]   & [ours] \\
          \midrule
        \addlinespace[.15em]
         {\multirow{2}{*}{$X_\mathrm{c}$}} & \textsc{tiny} & 6.5 $\pm$ 1.5  & 19.5 $\pm$ 3.7 & 8.8 $\pm$ 1.7 & 12.9 $\pm$ 0.9 & 6.9 $\pm$ 1.1 & \textbf{5.4 $\pm$ 0.5}  \\
          &  \textsc{small} & 6.5 $\pm$ 1.5   & 11.2 $\pm$ 1.0 & 6.4 $\pm$ 0.7 & 8.6 $\pm$ 1.5 & 4.4 $\pm$ 0.6 & \textbf{4.1 $\pm$ 0.5}  \\
            &  \textsc{medium} & 6.5 $\pm$ 1.5  & 9.7 $\pm$ 1.4 & 7.9 $\pm$ 1.3 & 8.9 $\pm$ 3.6 & 4.0 $\pm$ 1.0 & \textbf{3.9 $\pm$ 0.4}\\
        \addlinespace[.15em]
    \midrule
            \addlinespace[.15em]
         {\multirow{2}{*}{$X_\mathrm{CMU}$}} & \textsc{tiny} & 16.9 & 33.4 & 16.8 & 22.8 $\pm$ 1.5 & \textbf{14.3 $\pm$ 0.5} & 14.7 $\pm$ 0.8  \\
          &  \textsc{small} & 16.9   & 20.6 & 15.5 & 15.7 $\pm$ 0.2 &\textbf{ 12.6 $\pm$ 1.1} & 13.0 $\pm$ 0.3  \\
            &  \textsc{medium}  & 16.9   & 17.1 & 15.0 & 14.5 $\pm$ 0.3 &\textbf{ 11.1 $\pm$ 0.2}  & 11.9 $\pm$ 0.3  \\
        \addlinespace[.15em]
    \midrule
    \addlinespace[.15em]
             \multirow{2}{*}{$X_\mathrm{a}$} & \textsc{tiny} & \textbf{4.9 $\pm$ 0.7} & 76.1 $\pm$ 17.7 & 29.5 $\pm$ 8.8 & 40.9 $\pm$ 8.2 & 15.2 $\pm$ 6.9 & 9.5 $\pm$ 1.6  \\
           & \textsc{small} &  \textbf{4.9 $\pm$ 0.7} & 33.0 $\pm$ 5.8 & 10.6 $\pm$ 1.4 & 14.3 $\pm$ 3.1 & 6.8 $\pm$ 0.6 & 6.9 $\pm$ 0.9  \\
            & \textsc{medium} & \textbf{4.9 $\pm$ 0.7} & 27.4 $\pm$ 5.1 & 9.9 $\pm$ 1.5 & 14.9 $\pm$ 2.9 & 5.8 $\pm$ 0.5 &  5.7 $\pm$ 0.1 \\ 
         \addlinespace[.15em]
        \bottomrule
    \end{tabular}
}
    \caption{Average speaker-level WER for verbatim transcription on untuned and fine-tuned Whisper models. For each case study of children's speech or atypical adult speech, tuned and end-to-end models are respectively trained on $X_\mathrm{c}$ and $X_\mathrm{a} $. Averages and standard deviations are computed across three trials with different randomized splits. $X_\mathrm{CMU}$ is used only to evaluate models trained on $X_\mathrm{c}$; since the naive and untuned baselines do not require model training, $X_\mathrm{CMU}$ has no variation in these evaluations. }
    \label{tab:wers}
\end{table*}


Lastly, we evaluate models trained with $X_\mathrm{c}$ on the CMU Kids Corpus collected through the Linguistics Data Consortium~\cite{cmukids}, denoted by $X_\mathrm{CMU}$. This dataset consists of read speech from children aged 6 to 11 with over 5000 utterances in total. We re-annotated this dataset for a verbatim capture of miscues, and reserve this dataset to evaluate generalization performance on children's read speech. 

Data protocols are independently designed so there is no overlap in reading prompts or speakers between these three datasets; there are some common reading prompts across speakers within each dataset. For both case studies ($X_\mathrm{c}$ and $X_\mathrm{a}$), datasets are split into 70\% train, 10\% validation and 20\% test sets such that there is no speaker overlap. We conduct three trials where each trial consists of different random splits and report the average trial and standard deviation across trials.

\subsection{Evaluation}
\label{subsec:eval}

\begin{figure}[t]
    \includegraphics[width=\linewidth]{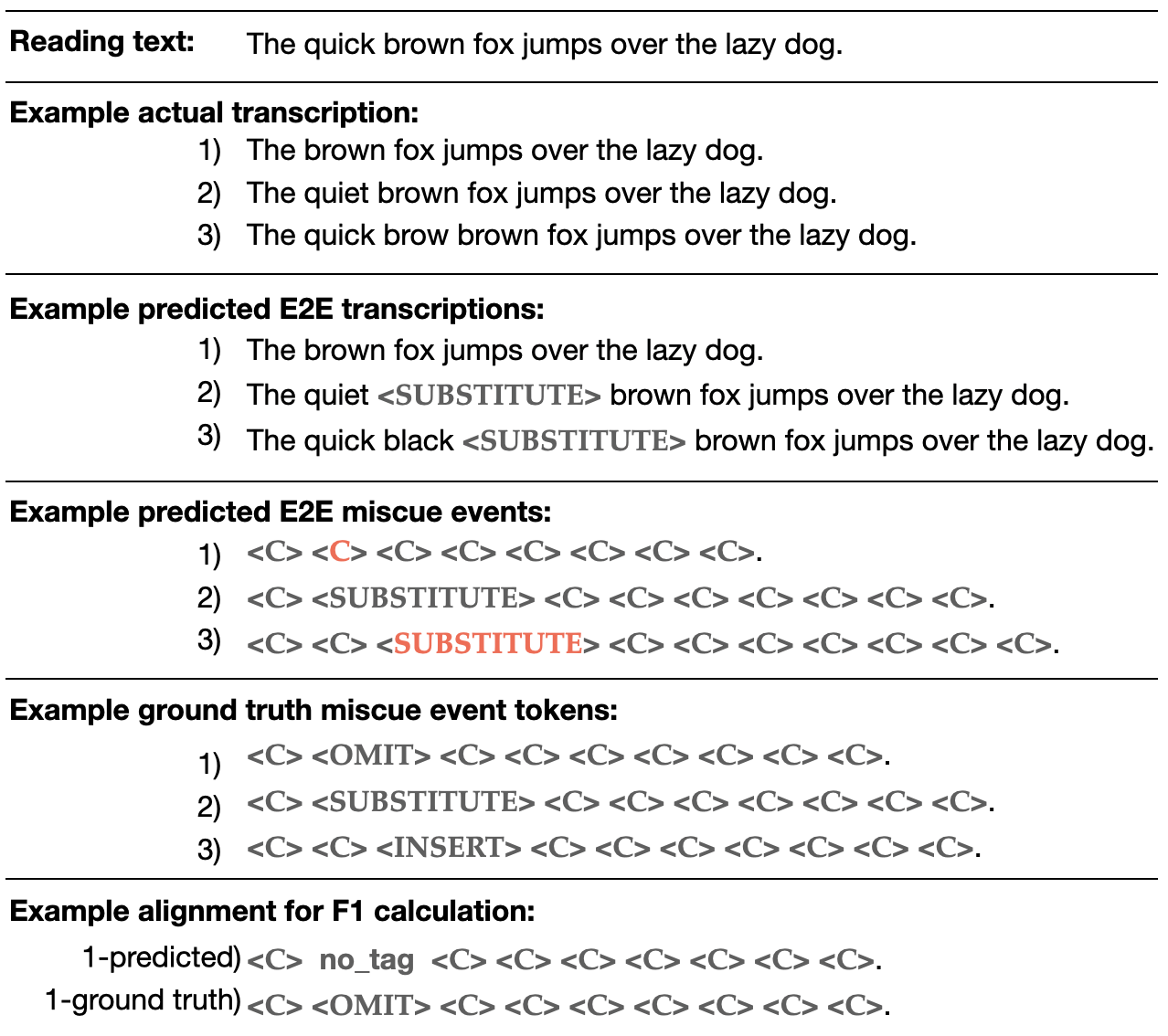}
    \caption{Examples of ground truth transcripts, predicted transcripts, and processing applied to evaluate MD performance given a reading text prompt. Predicted transcripts are processed in terms of predicted miscue event tokens and `\textsc{correct}' tokens. Note that `\textless\textsc{correct}\textgreater' is shown as `\textless\textsc{C}\textgreater' for readability and incorrectly predicted miscue event tokens are highlighted in red. F1 is computed using the predicted and ground truth miscue event tokens, with \texttt{no\_tag} inserted where needed for alignment.}
    \label{fig:miscue_wer_examples}
\end{figure}

We evaluate the E2E system performance for verbatim transcription and MD and compare to existing baselines. Baselines include untuned models with and without prompting to better understand the effect of prompting without any model fine-tuning. Additionally, we fine-tuned models with and without prompting to better understand the effect of tuning and the compounded effect of tuning with prompted context. Lastly, we consider a naive method that uses the reading text prompt as the predicted transcript; it cannot identify any miscues, but gives insights into the prevalence of miscues in our datasets. 

To evaluate verbatim transcription performance, we compute a speaker-level Word Error Rate (WER) using predicted transcripts and ground truth labels with miscue event tokens removed. To evaluate MD performance we use model output for E2E or post-hoc calculation for other setups to identify a miscue event for each word; words that are not associated with a miscue event (i.e., that are read correctly) are associated with `\textless\textsc{correct}\textgreater' tokens. We then align the predicted miscue event tokens with the ground truth miscue event tokens and compute F1 scores for each miscue event type. Note that F1 scores require all ground truth tokens have a corresponding aligned predicted token; any tokens that do not have a corresponding aligned token when computing alignment are given a corresponding \texttt{no\_tag} for the purpose of F1 calculation. We do not compute statistics on this \texttt{no\_tag} token itself. Figure \ref{fig:miscue_wer_examples} shows some examples of the processing applied for evaluation. 

\section{Results and Discussion}

We compare our approach to the verbatim transcription and MD baselines specified in Section \ref{subsec:eval} in Tables \ref{tab:wers} and \ref{tab:miscues_wer}, respectively.

\begin{table*}[h]
    \centering
    \resizebox{\linewidth}{!}{
    \begin{tabular}{lllcccccccc}
    \toprule
         & \textsc{Data} & \textsc{Miscue Type}  & \textsc{Naive}  & \multicolumn{2}{c}{\textsc{Untuned}}& 
        \multicolumn{2}{c}{\textsc{Tuned}} & \multicolumn{2}{c}{\textsc{End-to-End}} \\
        \cmidrule(lr){5-6}
        \cmidrule(lr){7-8}
        \cmidrule(lr){9-10}
        &  & & & \textsc{Unprompt} & \textsc{Prompt} & \textsc{Unprompt} & \textsc{Prompt} [ours]  & \textsc{Calc} [ours] & \textsc{Pred} [ours] \\
          \bottomrule
        \addlinespace[.15em]
       \parbox[c]{2mm}{\multirow{12}{*}{\rotatebox[origin=c]{90}{{\texttt{tiny.en }\hspace{0.1mm} } }}}  & {\multirow{2}{*}{$X_\mathrm{c}$}} 
        & \textless\textsc{substitute}\textgreater & .000 (.000) & .100 (.925)& .275 (.575)& .298 (.788)& .420 (.452)& \textbf{.472 (.439)}& .330 (.264)\\
        & & \textless\textsc{omit}\textgreater & .000 (.000) & .247 (.499)& .381 (.650)& .461 (.688)& \textbf{.556 (.620)}& .535 (.520)& .394 (.323)\\
        & & \textless\textsc{insert}\textgreater & .000 (.000) & .287 (.457)& .361 (.379)& .449 (.619)& .396 (.356)& \textbf{.505 (.444)}& .327 (.265)\\
        & & \textless\textsc{correct}\textgreater & .978 (1.00) & .752 (.606)& .963 (.939)& .956 (.921)& .982 (.983)& \textbf{.986 (.989)}& .981 (.992)\\
        
        \addlinespace[.15em]
            \cmidrule(lr){2-10}
            \addlinespace[.15em]
        & {\multirow{2}{*}{$X_\mathrm{CMU}$}} 
         & \textless\textsc{substitute}\textgreater & .000 (.000) & .178 (.883)& .389 (.526)& .390 (.794)& .407 (.388)& \textbf{.411 (.364)}& .303 (.247)\\
        & & \textless\textsc{omit}\textgreater & .000 (.000) & .306 (.678)& .627 (.883)& .513 (.862)& \textbf{.629 (.788)}& .540 (.619)& .342 (.322)\\
        & & \textless\textsc{insert}\textgreater & .000 (.000) & .317 (.393)& .430 (.362)& \textbf{.500 (.562)}& .416 (.337)& .474 (.395)& .334 (.275)\\
        & & \textless\textsc{correct}\textgreater & .943 (1.00) & .529 (.424)& .939 (.920)& .914 (.857)& .950 (.952)& \textbf{.952 (.963)}& .945 (.965)\\
        \addlinespace[.15em]
       \cmidrule(lr){2-10}
    \addlinespace[.15em]
       & {\multirow{2}{*}{$X_\mathrm{a}$}} 
        & \textless\textsc{substitute}\textgreater & .000 (.000) & .043 (.932)& .157 (.651)& .095 (.858)& .215 (.468)& \textbf{.241 (.468)}& .209 (.275)\\
        & & \textless\textsc{omit}\textgreater & .000 (.000) & .111 (.668)& .158 (.744)& .180 (.713)& .230 (.590)& \textbf{.267 (.519)}& .158 (.301)\\
        & & \textless\textsc{insert}\textgreater & .000 (.000) & .153 (.506)& .290 (.409)& .217 (.592)& .294 (.352)& \textbf{.389 (.471)}& .327 (.376)\\
        & & \textless\textsc{correct}\textgreater & \textbf{.990 (1.00)} & .584 (.431)& .922 (.872)& .847 (.760)& .977 (.965)& .982 (.976)& .975 (.982)\\

        \addlinespace[.15em]
          \bottomrule
        \addlinespace[.15em]
        \parbox[t]{2mm}{\multirow{12}{*}{\rotatebox[origin=c]{90}{{\texttt{medium.en}\hspace{4mm}} }}}   &  {\multirow{2}{*}{$X_\mathrm{c}$}} 
        & \textless\textsc{substitute}\textgreater & .000 (.000) & .108 (.889)& .388 (.516)& .423 (.736)& .548 (.510)& \textbf{.541 (.491)}& .519 (.470)\\
        & & \textless\textsc{omit}\textgreater & .000 (.000) & .289 (.510)& .393 (.610)& .564 (.685)& \textbf{.650 (.654)}& .604 (.575)& .518 (.471)\\
        & & \textless\textsc{insert}\textgreater & .000 (.000) & .304 (.335)& .332 (.292)& .585 (.678)& \textbf{.667 (64.9)}& .645 (.612)& .616 (.587)\\
        & & \textless\textsc{correct}\textgreater & .978 (1.00) & .787 (.652)& .977 (.969)& .977 (.962)& \textbf{.989 (.992)}& \textbf{.989 (.993)}& .988 (.992)\\
        \addlinespace[.15em]
       \cmidrule(lr){2-10}
            \addlinespace[.15em]
       &  {\multirow{2}{*}{$X_\mathrm{CMU}$}} 
       & \textless\textsc{substitute}\textgreater & .000 (.000) & .233 (.848)& .448 (.406)& \textbf{.481 (.715)}& \textbf{.481 (.404)}& .467 (.387)& .426 (.335)\\
        & & \textless\textsc{omit}\textgreater & .000 (.000) & .503 (.780)& .579 (.747)& .583 (.854)& \textbf{.628 (.768)}& .588 (.680)& .499 (.536)\\
        & & \textless\textsc{insert}\textgreater & .000 (.000) & .359 (.302)& .336 (.245)& .654 (.694)& \textbf{.671 (.657)}& .646 (.618)& .618 (.585)\\
        & & \textless\textsc{correct}\textgreater & .943 (1.00) & .797 (.670)& .952 (.955)& .944 (.915)& \textbf{.960 (.966)}& .954 (.963)& .950 (.962)\\
        \addlinespace[.15em]
       \cmidrule(lr){2-10}
    \addlinespace[.15em]
        & {\multirow{2}{*}{$X_\mathrm{a}$}} 
         & \textless\textsc{substitute}\textgreater & .000 (.000) & .048 (.917)& .215 (.629)& .213 (.753)& .286 (.543)& \textbf{.293 (.531)}& .254 (.445)\\
        & & \textless\textsc{omit}\textgreater & .000 (.000) & .127 (.732)& .228 (.714)& .299 (.741)& \textbf{.345 (.635)}& .322 (.586)& .189 (.368)\\
        & & \textless\textsc{insert}\textgreater & .000 (.000) & .241 (.419)& .301 (.377)& .443 (.664)& \textbf{.517 (.599)}& .498 (.579)& .435 (.556)\\
        & & \textless\textsc{correct}\textgreater & \textbf{.990 (1.00)} & .667 (.513)& .967 (.943)& .955 (.924)& .984 (.978)& .985 (.979)& .984 (.979)\\
        \addlinespace[.15em]
        \bottomrule
    \end{tabular}
    }
    \caption{Mean speaker-level F1 and recall (parenthesized; gives insight into F1 composition) for each miscue type, averaged across splits using \texttt{tiny.en} and \texttt{medium.en} setups. Highest F1 for each setup/miscue type is bolded. Due to limited space, \texttt{small.en} is not shown.}
    \label{tab:miscues_wer}
\end{table*}

\subsection{Verbatim transcription}
\label{ref:results_transcript}
Starting with untuned, unprompted performance, which represents the de facto application of ASR, we observe that WERs are much higher than 15\% for children and atypical speech using \texttt{tiny.en} models, making it unsuitable for most ASR applications~\cite{projecteuphonia}. Performance improves upon increasing model sizes, but is still insufficient for $X_\mathrm{CMU}$ and $X_\mathrm{a}$. We observe that $X_\mathrm{CMU}$ has a higher WER than $X_\mathrm{c}$, likely due to the higher rate of miscues observed in $X_\mathrm{CMU}$ than $X_\mathrm{c}$, indicated by the difference in WERs when naively predicting the prompted reading text. Notably, although this data shift in frequency of miscues between training and test data leads to challenges for good generalization performance, we observed our proposed approaches generally improving upon existing baselines.
Lastly, we observe that $X_\mathrm{a}$ has a very low overall miscue rate as naive prediction performs well for verbatim transcription. However this naive approach cannot detect miscues, whereas our proposed approaches using larger models achieve a good balance in both verbatim transcription and MD.

Overall tuning and prompting decreases WERs and variance thereof substantially across all methods and datasets. We also observed that despite differences between the target text prompts and ground truth verbatim transcriptions, prompting is more effective than fine-tuning without prompting, further supporting the use of prompting for providing context to verbatim transcription. Furthermore, compounding tuning with prompting is highly effective, leading to WERs lower than 15\% for most setups, except \texttt{tiny.en} models. Lastly, we observe that the E2E approach has better or similar verbatim transcription performance to tuned models with prompting, especially when train and test-sets are drawn from the same data distribution (i.e., $X_\mathrm{c}$ and $X_\mathrm{a}$) and for smaller ASR models (i.e., \texttt{tiny.en}
). This finding suggests the additional task of E2E MD may be helpful in improving verbatim transcriptions.  However for unseen data distributions (i.e. $X_\mathrm{CMU}$), tuning with prompting slightly outperforms E2E, suggesting more robustness against data shifts in miscue characteristics than E2E approaches.


\subsection{Miscue detection}
Using post-hoc miscue derivations, MD performance is correlated to verbatim transcription performance; untuned, unprompted ASR shows notably poor performance while tuning and prompting generally improve MD performance. 
Post-hoc calculated miscues in the E2E setup outperform E2E-predicted miscues, 
which generally had lower recall. 
The most common errors in the E2E setups falsely classify substitution miscues as correct, presenting opportunities for future improvements. 

For smaller models, miscues calculated post-hoc E2E verbatim transcripts outperform those from tuning with prompts. However, in larger models, tuning with prompts outperforms E2E approaches, suggesting smaller models benefit more from the additional task of miscue predictions than larger models. 
%


For children's speech, we observe F1 scores considerably higher than prior work, which showed a macro-F1 of 0.44 and 0.38 respectively for detecting an occurrence of a miscue and miscue classification on $X_\mathrm{CMU}$~\cite{venkatasubramaniam2023end}. However, there are key differences between this prior work and our own that prevent direct F1 comparisons, in that the prior work defined a different taxonomy of miscues, such as disregarding general miscue insertions but including a subclass of insertions pertaining to stuttering (i.e., repeated words or sounds). For atypical adult speech, $X_\mathrm{a}$, there is an overall lower performance compared to children's read speech. Observed  low precision is likely due to relative rarity and lower diversity of miscues observed; $X_\mathrm{a}$ is particularly imbalanced, where insertion errors are most common with substitution and omission errors being quite rare. 

\section{Conclusion}

Our experiments covered a range of approaches towards reading error annotation through modifying a large-scale speech foundation model. Extending prior methods that primarily focus on updating the language model or fine-tuning to improve recognition performance, we propose two novel strategies for training an E2E model to jointly predict verbatim transcriptions and miscue annotations: incorporating the reading text via prompting and extending the tokenizer vocabulary to include miscue detection tokens. We conducted two case studies on typical children's and atypical adult read-aloud speech to evaluate our approach along with state-of-the-art baselines and found that prompting with the reading text is highly effective in improving verbatim transcription performance and surprisingly more so than fine-tuning. Compounding fine-tuning with prompting further improves performance and the addition of miscue detection tokens also improves performance especially for smaller models. We also show that models can directly predict miscues in an end-to-end manner, although post-hoc calculation approaches are still most accurate. These findings fill gaps in miscue detection research where the target reading text is not always considered during transcription and highlights the potential of and need for further exploration into E2E approaches.

\bibliographystyle{IEEEtran}
\bibliography{mybib}

\end{document}